%% file: main_arxiv.tex
\documentclass[10pt,twocolumn,letterpaper]{article}

\usepackage[pagenumbers]{cvpr}              

\usepackage{graphicx}
\usepackage{amsmath}
\usepackage{amssymb}
\usepackage{booktabs}

\usepackage[pagebackref,breaklinks,colorlinks]{hyperref}

\usepackage{times}
\usepackage{epsfig}
\usepackage{graphicx}
\usepackage{amsmath, bm}
\usepackage{amssymb}
\usepackage{multirow}
\usepackage{microtype}
\usepackage[normalem]{ulem}
\useunder{\uline}{\ul}{}
\usepackage{color}
\usepackage{nicefrac}
\usepackage{comment}
\usepackage{enumitem}
\usepackage{adjustbox}
\usepackage{mathabx}
\usepackage{tabu}
\usepackage{booktabs}
\usepackage{diagbox}
\usepackage{cuted}
\usepackage{capt-of} 
\usepackage{multicol}
\usepackage[table,x11names]{xcolor}

\usepackage{xcolor,colortbl}
\definecolor{lgreen}{RGB}{236, 255, 201}

\usepackage[capitalize]{cleveref}
\crefname{section}{Sec.}{Secs.}
\Crefname{section}{Section}{Sections}
\Crefname{table}{Table}{Tables}
\crefname{table}{Tab.}{Tabs.}

\begin{document}

\title{\Large Seeing the Whole Picture: Distribution-Guided Data-Free Distillation\\ for Semantic Segmentation}

\author{Hongxuan Sun, Tao Wu$^*$ \vspace{1mm} \\ 
National University of Defense Technology\\
{\tt\small \{shx0227, wutao\}@nudt.edu.cn}
}
\maketitle
\def\thefootnote{$*$}\footnotetext{Corresponding author.}
\begin{abstract}
    Semantic segmentation requires a holistic understanding of the physical world, as it assigns semantic labels to spatially continuous and structurally coherent objects rather than to isolated pixels. However, existing data-free knowledge distillation (DFKD) methods—primarily designed for classification—often disregard this continuity, resulting in significant performance degradation when applied directly to segmentation tasks.

    In this paper, we introduce DFSS, a novel data-free distillation framework tailored for semantic segmentation. Unlike prior approaches that treat pixels independently, DFSS respects the structural and contextual continuity of real-world scenes. Our key insight is to leverage Batch Normalization (BN) statistics from a teacher model to guide Approximate Distribution Sampling (ADS), enabling the selection of data that better reflects the original training distribution—without relying on potentially misleading teacher predictions. Additionally, we propose Weighted Distribution Progressive Distillation (WDPD), which dynamically prioritizes reliable samples that are more closely aligned with the original data distribution early in training and gradually incorporates more challenging cases, mirroring the natural progression of learning in human perception. Extensive experiments on standard benchmarks demonstrate that DFSS consistently outperforms existing data-free distillation methods for semantic segmentation, achieving state-of-the-art results with significantly reduced reliance on auxiliary data.
\end{abstract}

\input{sec1_Introduction}
\input{sec2_related_work}
\input{sec3_method}
\input{sec4_experiments}

\section{Conclusion}
We propose DFSS, a data-free knowledge distillation framework specifically designed for semantic segmentation. DFSS selects open-world samples that are both task-relevant and semantically diverse, providing more informative and effective supervision for student training. To address the challenge of unreliable high-confidence predictions from the teacher, we introduce approximate distribution sampling, which aligns the selected data with the original feature distribution by leveraging the statistical cues embedded in batch normalization layers.

Furthermore, we present a weighted progressive distillation strategy that dynamically adjusts sample importance over training. This allows the student to focus on more reliable samples in the early stages and gradually incorporate harder examples, thereby improving generalization under challenging supervision.
Extensive experiments on multiple benchmarks demonstrate that DFSS consistently outperforms existing data-free distillation methods for semantic segmentation, achieving new state-of-the-art performance.

{\small
\bibliographystyle{ieee_fullname}
\bibliography{cite.bib}
}

\end{document}

%% file: sec1_Introduction.tex
\section{Introduction}
\label{sec:intro}

\begin{figure}[!t]
  \centering
  \includegraphics[width=\linewidth]{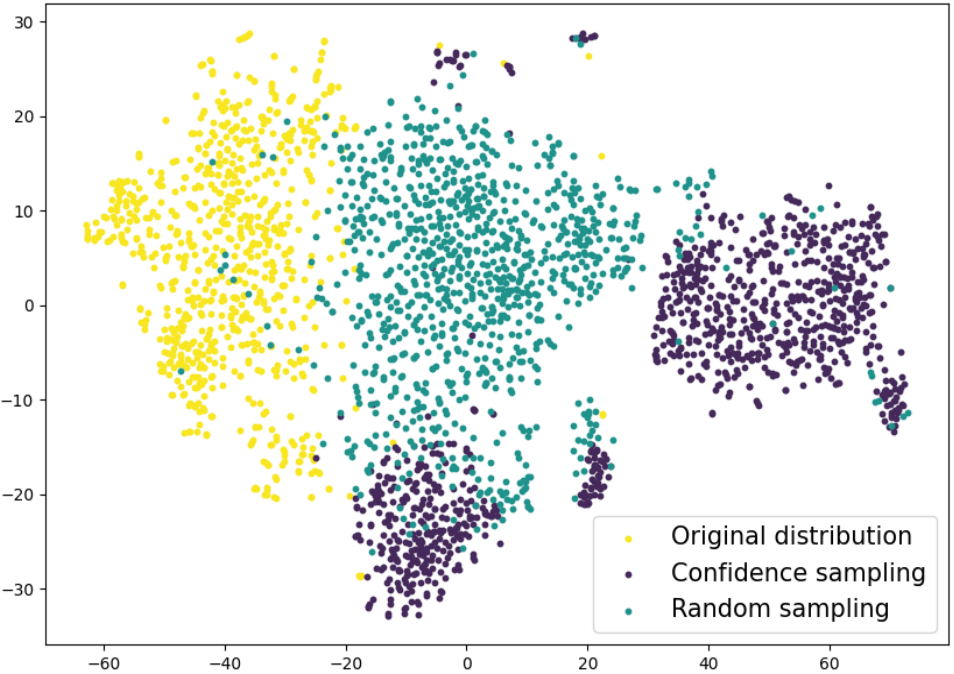}

  \caption{\textbf{t-SNE visualization of data distributions for different sampling strategies.} Compared to random sampling, confidence-based sampling leads to a more severe distribution shift from the original data, potentially introducing unintended biases that hinder model generalization.}
  \label{fig:mistakes}
\end{figure}

Deep neural networks (DNNs)\cite{alex2012alexnet,karen2015vggnet,szegedy2015googlenet} have achieved remarkable progress across a wide range of computer vision tasks\cite{he2016deep,lin2017feature,long2015fully}, from image classification to semantic segmentation. However, their deployment on resource-constrained devices remains challenging due to large model sizes, computational demands, and growing concerns over data privacy. To address these issues, model compression~\cite{lin2020pruning,tang2020scop,liu2017efficient,lin2020rotated,xu2019positiveunlabeled} and data-free techniques\cite{choi2020dfquantization,do2022momentumad,fang2021cmi,chen2019dafl,yin2020deepinv,kim2022naturalinv,luo2020largescaledf,fang2022fast} have emerged as promising directions, with recent work exploring open-world data sampling\cite{chen2021dfnd,tang2023kd3,wang2024odsd} to facilitate student learning without direct access to original training data.

While open-world sampling~\cite{chen2021dfnd,wang2024odsd} has shown strong results for classification, its direct application to semantic segmentation is far from trivial. Unlike classification, which assigns a single label per image, semantic segmentation requires dense, pixel-level predictions that capture the spatial continuity and object-level structure of real-world scenes. This fundamental difference reflects the physical world’s inherent continuity—semantic meaning arises not from isolated pixels, but from their spatial and contextual relationships. As a result, conventional sampling strategies that operate at the pixel or local patch level often disrupt these dependencies, leading to fragmented semantics and suboptimal knowledge transfer.

\begin{figure}[!t]
  \centering
  \includegraphics[width=\linewidth]{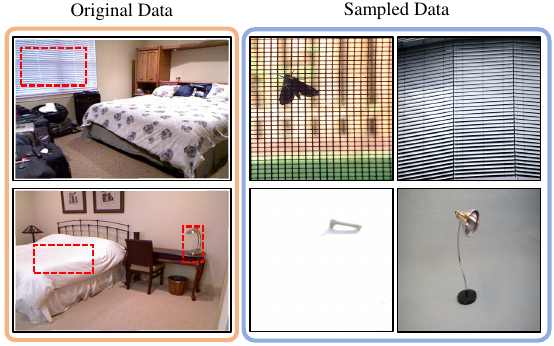}
  \caption{\textbf{Image visualization of confidence sampling results. }Segmentation networks are prone to overconfident predictions for images in the open-world data that share local features with the original data, resulting in a serious shift in the data distribution obtained by confidence sampling.}
  \label{fig:fake}
\end{figure}

Two major challenges underlie this problem. First, commonly used open-world datasets, such as ImageNet~\cite{deng2009imagenet}, lack the scene-level complexity and contextual richness required for effective segmentation. Second, as shown in Fig.~\ref{fig:fake}, confidence-based sampling methods~\cite{chen2021dfnd,wang2024odsd} tend to favor visually simple yet contextually uninformative samples, as teacher networks often assign high confidence to inputs that share superficial features with training images but lack meaningful structure. Besides, Fig.~\ref{fig:mistakes} shows that our analysis reveals that such approaches can cause greater distribution shift than even random sampling, ultimately impeding student generalization.

Motivated by the need for structure-aware sampling, we propose DFSS (Data-Free Knowledge Distillation for Semantic Segmentation), a framework that explicitly accounts for the physical and semantic continuity of the real world. DFSS first employs Approximate Distribution Sampling (ADS), harnessing Batch Normalization statistics from the teacher network to select data that aligns with the original distribution—circumventing the pitfalls of confidence-based selection. To further enhance learning, we introduce Weighted Distributed Progressive Distillation (WDPD) that adaptively adjusts sample importance: the student initially focuses on reliable examples that better resemble the original data distribution, and later incorporates harder, more complex cases, emulating the progressive nature of human learning. Our main contributions are summarized as follows:

\begin{itemize}
    \item We identify the unique challenges of data-free distillation for semantic segmentation and propose DFSS, a structure-aware framework that leverages BN statistics for principled data sampling, reducing reliance on unreliable teacher predictions.
    \item We introduce a progressive weighted distillation scheme that dynamically balances core and challenging examples, improving the robustness and diversity of student learning.
    \item Comprehensive experiments across multiple benchmarks demonstrate that DFSS achieves state-of-the-art performance, significantly reducing the need for auxiliary data while preserving spatial and contextual integrity.
\end{itemize}

\begin{figure*}[!t]
  \centering
  \includegraphics[width=\linewidth]
  {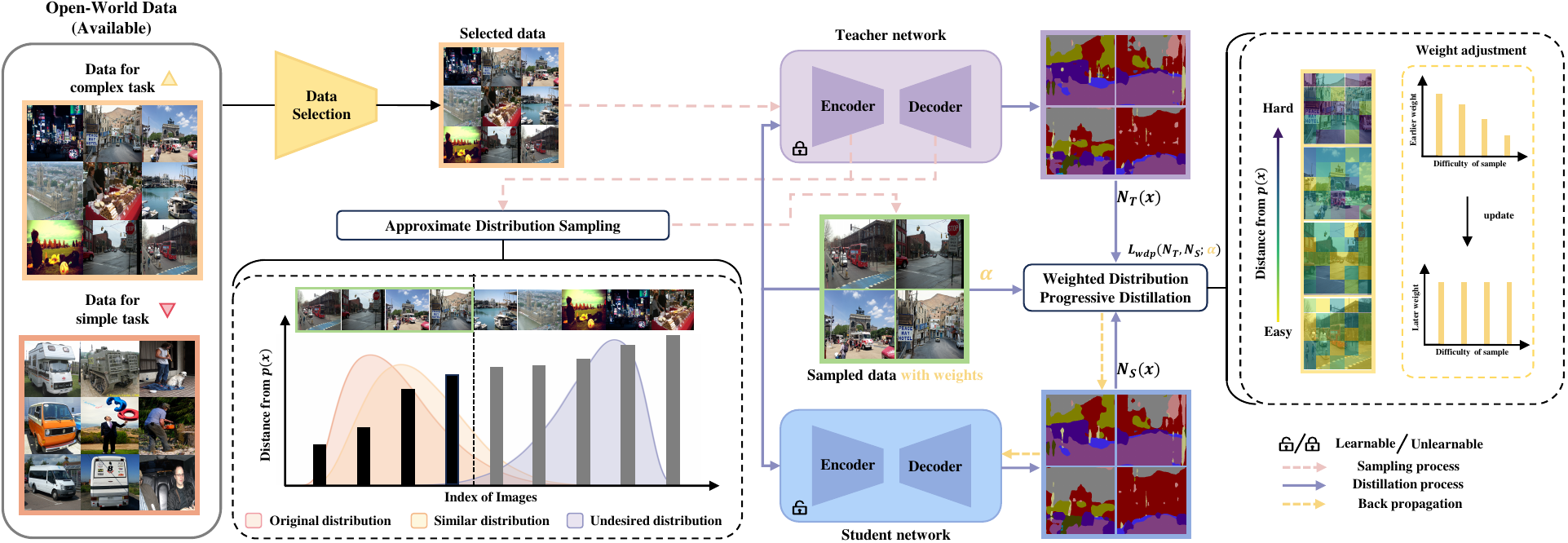}
  \caption{The pipeline of our proposed DFSS. Initially, we emphasize the selection of open-world data based on its relevance to semantic segmentation tasks during the data collection stage. Subsequently, the gathered unlabeled data undergoes approximate distribution sampling, which guarantees that the substitute dataset's distribution closely approximates the original data distribution while assigning appropriate weights to each sample based on its distribution characteristics. Ultimately, utilizing these weighted samples, the student network progressively acquires knowledge from challenging samples through our proposed weighted distribution progressive distillation strategy, facilitating more efficient and effective knowledge transfer.}
  \label{fig:framework}
\end{figure*}

%% file: sec2_related_work.tex
\section{Related Work}

\subsection{Data-Driven Knowledge Distillation for Semantic Segmentation}
Knowledge distillation has been widely investigated in the domain of image classification, with subsequent research endeavors aiming to extend distillation methodologies to more sophisticated dense prediction tasks, particularly semantic segmentation~\cite{he2019adaptation, liu2021inter, xie2018improvingfs, shu2021channel}. Nevertheless, the direct application of classification-oriented distillation approaches to dense prediction tasks frequently yields suboptimal performance. Consequently, researchers have developed more advanced distillation paradigms, including prototype-based distillation~\cite{wang2020intra} and structured adversarial distillation~\cite{liu2019structured}.

While these methods have demonstrated effectiveness, they necessitate meticulous feature extraction processes during knowledge transfer. Consequently, recent research has concentrated on developing transformation modules for adaptive feature extraction. AttnFD~\cite{mansourian2024attnfd} enhances feature representation through a convolutional attention mechanism that simultaneously captures channel-wise and spatial information. NFD~\cite{liu2022nfd} employs feature normalization to eliminate magnitude information during distillation, thereby preventing the student network from merely replicating the teacher's feature response magnitudes. To address feature semantic structure discrepancies, MGD~\cite{yue2020mgd} leverages the Hungarian matching algorithm to align the student's intermediate feature maps with the most semantically similar counterparts from the teacher model, facilitating more effective knowledge acquisition.

While these methods produce compressed models with competitive performance, their reliance on original training data poses practical challenges. Privacy regulations and data transmission constraints often render the training data inaccessible. This fundamental limitation has spurred the development of data-free model compression.

\subsection{Data-Free Knowledge Distillation}
Recent research has made significant progress in DFKD, with existing approaches primarily categorized into two distinct paradigms based on their alternative data sources: generation-based methods and sampling-based methods.\\
\textbf{Generation-based Methods.} Generation-based approaches synthesize alternative data through generative modules. DAFL~\cite{chen2019dafl} pioneered the use of generative networks for DFKD, establishing an architectural foundation for subsequent research. To enhance data diversity, CMI~\cite{fang2021cmi} introduces a contrastive model inversion technique that constrains the generated data distribution to better approximate real data characteristics. SpaceshipNet~\cite{yu2023spaceshipnet} advances this paradigm by implementing channel-level feature map mixing and incorporating spatial activation constraints, thereby encouraging the student network to focus on the teacher's activated regions. DFAD~\cite{fang2019dfad} further extends this direction by developing an adversarial distillation framework that estimates and optimizes an upper bound on the challenging-to-handle model discrepancies between teacher and student networks.
Owing to its generalizability and model-agnostic difference metrics, DFAD represents the pioneering DFKD framework applicable to semantic segmentation tasks, subsequently inspiring numerous extensions in this domain. To address computational efficiency concerns, recent research has focused on optimizing performance with reduced synthetic data requirements. Fast~\cite{fang2022fast} introduces a meta-generator approach that extracts common features for initialization, facilitating rapid generation of diverse data instances. SSD-KD~\cite{liu2024ssd} advances this direction by implementing a dynamic replay buffer coupled with a reinforcement learning strategy, achieving balanced synthetic sample generation and enabling effective distillation with minimal synthetic data scales.
However, generation-based DFKD approaches inevitably incur substantial computational overhead through their generative modules, and despite advancements in generation quality, synthetic samples inherently lack the rich implicit information and natural variations present in real-world data.\\
\textbf{Sampling-based Methods.} Sampling-based approaches facilitate distillation by leveraging open-source unlabeled data as alternative training samples. Given the inherent variability in data quality, the development of reasonable sampling strategies becomes crucial for effective data selection. DFND~\cite{chen2021dfnd} implements a rigorous confidence-based filtering mechanism coupled with a learnable noise-adaptation matrix to mitigate label noise generated by the teacher network on unlabeled images. To tackle distribution shift challenges, KD$^3$~\cite{tang2023kd3} introduces a dual-network dynamic instance sampling strategy, complemented by network sharing and contrastive learning mechanisms to enhance teacher-student alignment in the feature space. Addressing domain shift issues, ODSD~\cite{wang2024odsd} develops an adaptive prototype sampling technique for optimal data selection, while incorporating contrastive learning to establish robust structured relationships that effectively suppress label noise.
While effective for semantic segmentation, current methods primarily adapt classification-oriented DFKD, overlooking segmentation's unique data requirements. This leads to feature complexity mismatches between collected and original data. Moreover, the prevalent confidence-based sampling approach proves inadequate, as open-world data often contains out-of-distribution samples that trigger misleading high-confidence predictions, ultimately degrading the student's performance.

%% file: sec3_method.tex
\section{Method}
\label{sec:method}
In this section, we first briefly discuss the background, and then introduce our methods.
\subsection{Background}
Knowledge distillation~\cite{hiton2015distill} has become a prevalent approach for model compression. The framework comprises two phases: training a sophisticated teacher model $N_{T}$, and transferring its knowledge to a compact student model $N_{S}$. Given training data $\mathcal{D}\!=\!\left\{\left(x_i, y_i\right)\right\}_{i=1}^{|\mathcal{D}|}$, where $|\cdot|$ is the data cardinality, $N_{S}$ can be distilled by:
\begin{equation}
    \label{eq:kd}
    L(N_{S})=H_{task}(N_{S}(x),y)+\lambda H_{kd}(N_{S}(x),N_{T}(x)),
\end{equation}
where $H_{task}$ denotes the task-specific loss function that minimizes the discrepancy between the predictions of $N_{S}$ and ground-truth labels; $H_{kd}$ signifies the knowledge distillation loss that facilitates the transfer of knowledge from $N_{T}$ to $N_{S}$; $\lambda$ represents the trade-off parameter.

From a data distribution perspective, knowledge distillation seeks to minimize the performance discrepancy between the teacher and student networks under the original data distribution. Let $D(N_{T},N_{S})$ represent this performance gap, if $\mathcal{D}$ follows data distribution $p$, the optimization objective can be formally expressed as:
\begin{equation}
    \label{eq:kd_goal}
    \min_{N_{S}}D(N_{T},N_{S};p).
\end{equation}

In DFKD scenarios, $\mathcal{D}$ required by conventional knowledge distillation methods is typically inaccessible. Instead, we rely on unlabeled data collected from the open world, denoted as $\bar{\mathcal{D}}\!=\!\left\{\bar{x}_i\right\}_{i=1}^{|\bar{\mathcal{D}}|}$. This limitation necessitates the reformulation of the loss function as follows:
\begin{equation}
    \label{eq:dfkd}
    L(N_{S})=H_{kd}(N_{S}(\bar{x}),N_{T}(\bar{x})).
\end{equation}

The optimization objective in Eq.~\ref{eq:kd_goal} thus becomes:
\begin{equation}
    \label{eq:kfkd_goal}
    \min_{N_{S}}D(N_{T},N_{S};\bar{p}),
\end{equation}

\noindent where $\bar{p}$ is the distribution of $\mathcal{\bar{D}}$. However, direct distillation using $\mathcal{\bar{D}}$ faces two fundamental challenges: the scale disparity ($|\mathcal{D}| \ll |\mathcal{\bar{D}}|$) and the distribution shift ($p \ne \bar{p}$). These issues are particularly evident in the domain gap of image complexity between $\mathcal{D}$ and $\mathcal{\bar{D}}$, which significantly affects distillation performance. Consequently, it becomes imperative to implement a selective sampling strategy from $\bar{\mathcal{D}}$, aiming to achieve efficient knowledge distillation through minimal yet maximally informative samples that effectively leverage the teacher's implicit knowledge.

\subsection{Collected Data Selection Principles}
Current sampling-based distillation methods typically follow a two-stage process: sampling a subset $\mathcal{\widehat{D}} \subset \mathcal{\bar{D}}$ followed by distillation using $\mathcal{\widehat{D}}$. However, these approaches often underestimate the importance of the initial data collection phase, which fundamentally determines the student's performance lower bound. We propose two key principles for open-world data selection: \textbf{task relevance} and \textbf{data abundance}. Formally, given the task $\mathcal{T}$ associated with original dataset $\mathcal{D}$ and $\bar{\mathcal{T}}$ associated with collected dataset $\mathcal{\bar{D}}$, we need to satisfy the following relationships at the same time:
\begin{equation}
\label{eq:select}
\mathcal{T} \approx \bar{\mathcal{T}}, \quad c(\mathcal{D}) \ll c(\bar{\mathcal{D}}),
\end{equation}

\noindent where $c(\cdot)$ measures dataset complexity. The first term in Eq.~\ref{eq:select} ensures $\bar{\mathcal{T}}$ maintains task similarity with ${\mathcal{T}}$, guaranteeing sufficient alignment in task characteristics. The last term is evaluated from two perspectives: data cardinality and information richness:
\begin{equation}
\label{eq:data}
    |\mathcal{D}| \ll |\mathcal{\bar{D}}|, \quad R(\mathcal{D}) \le R(\mathcal{\bar{D}}),
\end{equation}

\noindent where $R(\cdot)$ quantifies the information richness of datasets through image information entropy~\cite{shannon1948information} statistics computed across all images.

The selection criteria in Eq.~\ref{eq:data} are theoretically grounded in two aspects: sufficient data scale increases the probability of sampling instances that approximate the original data distribution, while high information richness ensures comprehensive feature representation. Notably, as characteristics of $\mathcal{D}$ are unknown, $\mathcal{\bar{D}}$ should maximize both scale and information richness to better satisfy these requirements.

\begin{table*}[htbp]
\centering
\begin{tabular}{cccccc}
\hline
\textbf{Algorithm}     & \textbf{Required data} & \textbf{FLOPS} & \textbf{\#params} & \textbf{NYUv2} & \textbf{CamVid} \\ \hline
Teacher                & Original data          & 41.0G          & 24M               & 0.519          & 0.594           \\
Knowledge Distillation & Original data          & 5.54G          & 3.4M              & 0.380          & 0.535           \\ \hline
DAFL~\cite{chen2019dafl}                   & Generated data         & 5.54G          & 3.4M              & 0.105          & 0.010           \\
Fast~\cite{fang2022fast}                   & Generated data         & 5.54G          & 3.4M              & 0.366          & -               \\
SSD-KD~\cite{liu2024ssd}                 & Generated data         & 5.54G          & 3.4M              & 0.384          & -               \\ 
DFAD~\cite{fang2019dfad}                   & Generated data         & 5.54G          & 3.4M              & 0.364          & 0.535 \\ \hline
DFND~\cite{chen2021dfnd}                   & Unlabeled data         & 5.54G          & 3.4M              & 0.378          & 0.507{\dag}  \\
ODSD~\cite{wang2024odsd}                   & Unlabeled data         & 5.54G          & 3.4M              & 0.397         & -               \\ 
DFSS($\epsilon$=10K)    & Unlabeled data         & 5.54G          & 3.4M              & 0.492              & 0.579  \\
DFSS($\epsilon$=15K)    & Unlabeled data         & 5.54G          & 3.4M              & \underline{0.498}  & \underline{0.586}      \\
DFSS($\epsilon$=20K)    & Unlabeled data         & 5.54G          & 3.4M              & \textbf{0.506}  &\textbf{0.590}      \\ \hline
\end{tabular}
\caption{Performance of student on CamVid and NYUv2 datasets. \textbf{Bold} and \underline{underline} numbers denote the best and the second best results. \dag is the result of our reproduction. Since DFND does not specify the exact experimental setup for the semantic segmentation task, we reproduced it based on the sampling method described in the original paper. We set the sample size to 20K, and the rest of the experimental settings are identical to those in DFSS.}
\label{tab:sota}
\end{table*}

\subsection{Approximate Distribution Sampling}
As shown in Eq.~\ref{eq:dfkd}, when $\mathcal{D}$ is unavailable, the knowledge transfer relies entirely on $N_{T}(\widehat{x})$ as the supervision signal. Since $N_{T}$ is trained on $\mathcal{D}$, its knowledge is inherently bounded by $p(x)$. To ensure reliable predictions from $N_{T}$, the sampled data distribution $p(\widehat{x})$ must closely approximate $p(x)$. With $p(x)$ is fixed, our objective reduces to sampling $\mathcal{\widehat{D}}$ that minimizes the distribution discrepancy between $p(\widehat{x})$ and $p(x)$, which can be expressed as:
\begin{equation}
    \label{eq:dfkd_goal}
    \min_{\widehat{\mathcal{D}}}D(p(x),p(\widehat{x})).
\end{equation}

The open world contains abundant out-of-distribution data that may mislead segmentation networks into generating unreliable high-confidence predictions, ultimately compromising knowledge transfer effectiveness to the student. To address this challenge, we propose a simple yet effective sampling strategy that exploits the intrinsic properties of Batch Normalization (BN) layers in $N_{T}$. BN layers naturally preserve running statistics ($\mu_{bn},\sigma^{2}_{bn}$) that encode the original data distribution, these running statistics approximate the expected mean and variance of the original training data distribution. Therefore, when a sample passes through $N_{T}$'s feature extraction, the closer alignment between its feature map statistics (mean and variance) and BN running statistics ($\mu_{bn},\sigma^{2}_{bn}$) indicates the higher similarity to the original data distribution. Our objective thus becomes maintaining close consistency between sampled data's feature statistics and BN running statistics, formally expressed as:
\begin{equation}
\label{eq:minn}
\min_{\mathcal{\widehat{D}}}D(\mu(\widehat{x}),\mu_{bn})+D(\sigma^{2}(\widehat{x}),\sigma^{2}_{bn}),
\end{equation}
where $\mu(\widehat{x})$ and $\sigma^{2}(\widehat{x})$ represent the mean and variance of the feature map associated with sample $\widehat{x}$.

Finally, leveraging Eq.~\ref{eq:minn}, we establish the sampling metric by quantifying the discrepancy $d$:
\begin{equation}
\label{eq:cal_dis}
\begin{aligned}
    d=||\mu(\widehat{x})-\mu_{bn}||_{2}+
          ||\sigma^{2}(\widehat{x})-\sigma_{bn}^{2}||_{2}.
\end{aligned}
\end{equation}

Samples with smaller $d$ values are sampled for subsequent distillation.
Given that segmentation network's predictions are inherently vulnerable to disturbances, confidence scores cannot reliably indicate a sample's proximity to $p(x)$. Consequently, rather than designing intricate filtering mechanisms for network predictions, we propose a more effective strategy that directly utilizes the implicit distribution information encoded in BN layers for sampling.\\

\subsection{Weighted Distributed Progressive Distillation}  
\textbf{Weighted Distribution Distillation.} Using Eq.~\ref{eq:cal_dis}, we compute the distribution distance $d$ for each sample in $\mathcal{\widehat{D}}$. Samples with smaller $d$ values are closer to $p(x)$, enabling $N_{T}$ to provide more reliable supervision for training $N_{S}$. Therefore, our loss function explicitly incorporates these distribution characteristics, assigning higher weights to samples that better approximate $p(x)$:
\begin{equation}
\label{eq:wd}
    L_{wd}=\frac{1}{|\widehat{\mathcal{D}}|}\sum_{i=1}^{|\widehat{\mathcal{D}}|} \omega_{i}H_{l1}(N_{S}(\widehat{x_{i}}),N_{T}(\widehat{x_{i}})),
\end{equation}
where $\omega_{i}$ is the weight of the $i$-th sample, calculated by $d_{i}$:
\begin{equation}
\label{eq:weight}
    \omega_{i} =1-\frac{d_{i}-min\{d\}}{max\{d\}-min\{d\}}.
\end{equation}

The alignment quality between a sample and the original distribution determines the precision of the teacher's knowledge. Thus, we assign higher weights to better-aligned samples, allowing the student model to focus on more reliable supervision during training.\\
\textbf{Weighted Distribution Progressive Distillation.} While samples deviating from the original distribution may contain unreliable predictions, they still carry valuable information. Inspired by human learning progression, we propose a dynamic learning strategy where the student initially focuses on reliable samples to build robust feature representations, then gradually incorporates more challenging samples to enhance its generalization capability. This is achieved through a dynamic weighting mechanism that progressively increases the importance of harder samples during training:
\begin{equation}
\label{eq:wdp}
    L_{wdp}(t)=\frac{1}{|\widehat{\mathcal{D}}|}\sum_{i=1}^{|\widehat{\mathcal{D}}|} \alpha_{i}(t)H_{l1}(N_{S}(\widehat{x_{i}}),N_{T}(\widehat{x_{i}})),
\end{equation}
where $\alpha_{i}$ is the optimized weight factor of the $i$-th sample, dynamically adjusted by the following function:
\begin{equation}
\label{eq:weight_update}
\alpha_{i}(t)=\left\{\begin{array}{cl}
\omega_{i}+\frac{1-\omega_{i}}{I/2}t, & t \leq {I/2}, \\
1, & t> {I/2},
\end{array}\right.
\end{equation}

\noindent where $I$ represents the total number of iterations in training student network. Through Eq.~\ref{eq:wdp}, $N_{S}$ progressively transitions from targeted learning to balanced learning throughout the training process, thereby achieving enhanced performance.

\begin{figure*}[!t]
  \centering
  \includegraphics[width=\linewidth]
  {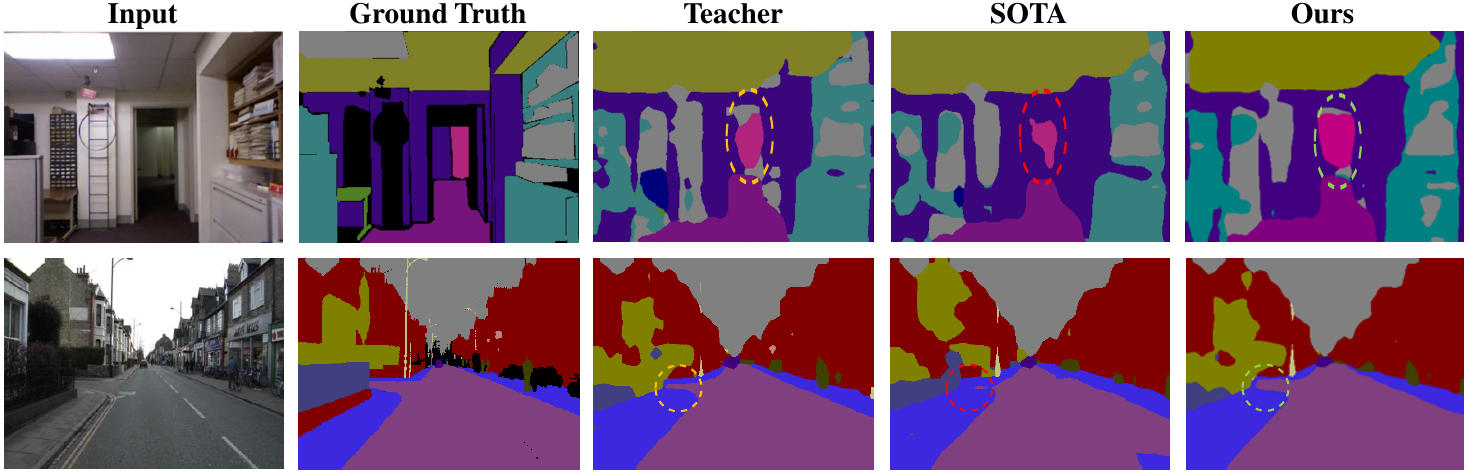}
  \caption{Segmentation results on NYUv2 and CamVid. The first row shows the visualization for the NYUv2 dataset, with the SOTA method being ODSD. The second row shows the visualization for the CamVid dataset, with the SOTA method being DFAD.}
  \label{fig:vis}
\end{figure*}

%% file: sec4_experiments.tex
\section{Experiments}
\subsection{Experimental Settings}
\noindent\textbf{Datasets.} We conducted comprehensive evaluations of DFSS using established benchmark datasets. Specifically, we assessed the proposed method's performance on the NYUv2~\cite{nathan2012nyuv2} and CamVid~\cite{gabriel2008camvid,gabriel2008camvid_video} datasets. Furthermore, guided by our proposed data selection principles, we selected the COCO~\cite{lin2015coco} dataset as the source for sampling, ensuring alignment with our task-driven and massive data requirements.\\
\textbf{Implementation Details.} Our implementation is based on PyTorch, with all experiments conducted on RTX 3090 GPUs. Following established practices, we employ DeeplabV3~\cite{chen2017deeplabv3} as our model architecture, utilizing ResNet50~\cite{he2016deep} as the teacher backbone and MobileNetV2~\cite{howard2017mobilenet} for the student network. We evaluate three sampling scales ($\epsilon$=10K, 15K, 20K), training for 150 epochs with SGD, batch size=192, and cosine learning rate scheduling.\\
\textbf{Baselines.} We compare our approach against two kinds of DFKD methods. The first kind involves methods that incur additional computational costs to generate data through dedicated generation modules, including DAFL~\cite{chen2019dafl}, DFAD~\cite{fang2019dfad}, Fast~\cite{fang2022fast}, and SSD-KD~\cite{liu2024ssd}. The second kind consists of sampling-based approaches that utilize readily available open-source unlabeled datasets, including DFND~\cite{chen2021dfnd} and ODSD~\cite{wang2024odsd}. Furthermore, we include a comparison with student models trained using the original training data as an upper-bound reference.

\subsection{Experimental Results}
\noindent\textbf{Comparison with SOTA DFKD Methods.} Table~\ref{tab:sota} shows the results achieved by our method and several SOTA methods. Through the implementation of our proposed DFSS, we achieve substantial improvements in DFKD performance for semantic segmentation tasks. Notably, our method surpasses current state-of-the-art performance even at $\epsilon$=10K, with the student's performance demonstrating consistent enhancement as $\epsilon$ increases. 
Furthermore, to provide more intuitive evidence, we present visual comparisons of segmentation results across different methods, as illustrated in Fig.~\ref{fig:vis}. The "Input" and "Ground Truth" columns display the test images and their corresponding true labels, respectively. "SOTA" represents the predictions generated by state-of-the-art methods on various datasets. The visualization results demonstrate that our method achieves superior alignment with the teacher network's predictions.

\begin{table}[h]
\centering
\begin{tabular}{ccc}
\hline
Method                 & Training Data Scale     & mIOU           \\ \hline
Teacher                & NYUv2             & 0.519          \\
KD                     & NYUv2             & 0.380          \\ \hline
DFAL                   & 960K (synthetic)        & 0.105          \\
DFAD                   & 960K (synthetic)        & 0.364          \\
Fast                   & 17K (
)         & 0.366          \\
SSD-KD                 & {\underline{16K} (synthetic)}   & 0.384          \\ \hline
ODSD                   & 200K (sampling)         & {\underline{0.397}}    \\
\textbf{DFSS}          & \textbf{10K} (sampling) & \textbf{0.492} \\ \hline
\end{tabular}
\caption{Performance comparison of various methods in terms of the amount of training data required.}
\label{tab:scale}
\end{table}

\noindent\textbf{Comparison with Training Data Scale.} We conduct a comparative analysis of training data requirements for existing methods on NYUv2. We focus on the actual data utilized in distillation, as shown in Table~\ref{tab:scale}, current generation-based methods have reduced synthetic data requirements through meta-learning techniques and the integration of dynamic replay buffers with reinforcement learning strategies. While existing sampling-based methods necessitate larger data volumes for training, they achieve superior student performance owing to the richer information inherent in real images. Notably, our method attains optimal performance while using a small amount of training data.

\begin{table}[h]
\centering
\resizebox{0.48\textwidth}{!}{
\begin{tabular}{@{}ccccccc@{}}
\toprule
                  & \multicolumn{3}{c}{NYUV2}      & \multicolumn{3}{c}{CamVid}     \\ \midrule
Collected Dataset & CS    & RS    & ADS            & CS    & RS    & ADS            \\
ImageNet          & 0.161 & 0.390 & 0.474          & 0.436 & 0.523 & 0.570          \\
COCO              & 0.388 & 0.431 & \textbf{0.483} & 0.539 & 0.545 & \textbf{0.578} \\
\bottomrule
\end{tabular}
}
\caption{Effectiveness of different collected datasets. RS and CS represent random sampling and confidence sampling, respectively.}
\label{tab:datasets}
\end{table}

\subsection{Ablation Studies}
\noindent\textbf{Effect of Different Sampling Strategies.} We use two datasets for sampling: ImageNet, which is commonly used in existing work, and COCO, which adheres to our proposed data selection principles. The experiments involved random sampling, confidence-based sampling, and approximate distribution sampling (ADS), while the distillation process was kept consistent without weighted distribution progressive distillation (WDPD). Table~\ref{tab:datasets} summarizes the results, the findings show that random sampling on COCO easily surpasses the SOTA methods (0.397 on NYUv2 and 0.535 on CamVid), while on ImageNet only ADS can exceed SOTA, although its effectiveness is consistently weaker compared to ADS on COCO. Furthermore, the poor results obtained from confidence-based sampling on both datasets highlight fundamental issues with existing sampling methods.
\begin{figure}[!t]
  \centering
  \includegraphics[width=\linewidth]{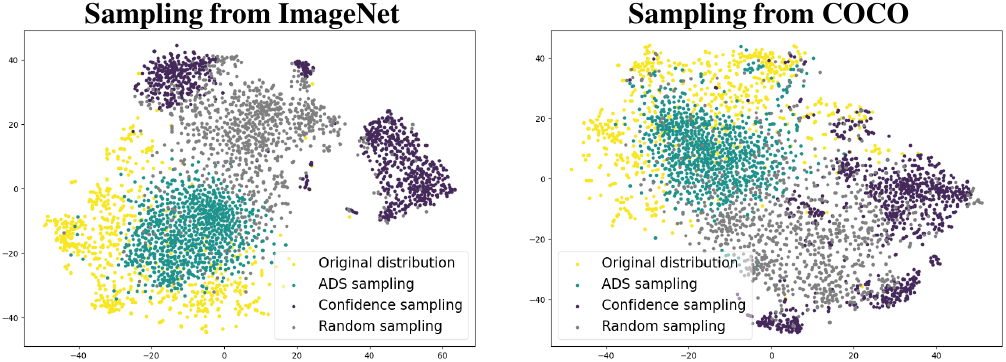}
  \caption{t-SNE visualization of the data distributions on NYUv2. The yellow points represent the original domain data, while the green, purple, and gray points correspond to the data obtained by three sampling methods: ADS, confidence sampling, and random sampling.}
  \label{fig:tsne-nyuv2}
\end{figure}

\begin{figure}[!t]
  \centering
  \includegraphics[width=\linewidth]{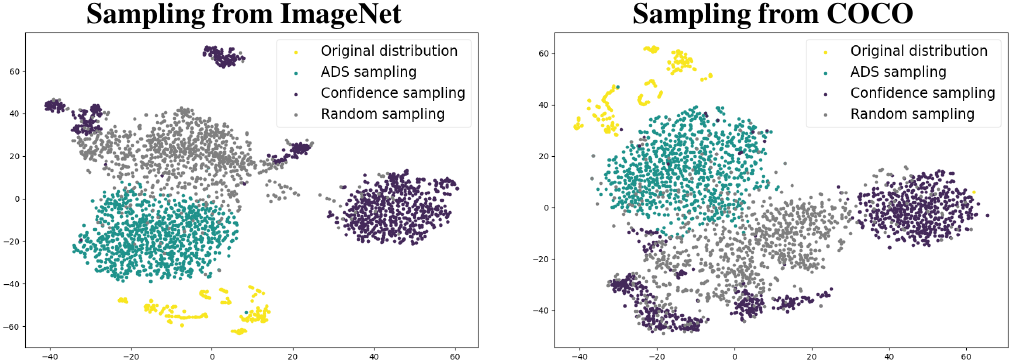}
  \caption{t-SNE visualization of the data distributions on CamVid. The yellow points represent the original domain data, while the green, purple, and gray points correspond to the data obtained by three sampling methods: ADS, confidence sampling, and random sampling.}
  \label{fig:tsne-camvid}
\end{figure}

\noindent\textbf{Visualization.} We establish baseline comparisons through random sampling visualizations. For each segmentation task, we apply different sampling methods to extract 1K samples from COCO and ImageNet. Fig.~\ref{fig:tsne-nyuv2} shows that ADS achieves superior alignment with the NYUv2 data distribution compared to other methods. Although ADS shows limited alignment for CamVid in Fig.~\ref{fig:tsne-camvid}, it still outperforms both random and confidence sampling in distribution proximity. 

\noindent\textbf{Effect of Different Distillation Strategies.} In Table~\ref{tab:distill}, we present a comprehensive ablation study to evaluate our distillation strategies. Using vanilla distillation as a baseline, we examine the impact of Weighted Distribution Distillation (WDD) and WDPD on the performance of the student model. Although WDD demonstrates measurable improvements, it still underperforms compared to vanilla distillation, suggesting that lower weight samples, while providing unreliable predictions, still retain valuable knowledge. When employing the WDPD, the student model initially develops robust feature extraction capabilities through high-quality samples, which subsequently enables more effective learning from challenging samples in later training phases.

\begin{table}[h]
\centering
\resizebox{0.475\textwidth}{!}{
\begin{tabular}{@{}p{1.5cm}cccccc@{}}
\toprule
       & \multicolumn{3}{c}{NYUv2}         & \multicolumn{3}{c}{CamVid}        \\ \midrule
Method & Baseline & +WDD  & +WDPD          & Baseline & +WDD  & +WDPD          \\
mIOU   & 0.483    & 0.455 & \textbf{0.492} & 0.578    & 0.563 & \textbf{0.579} \\ \bottomrule
\end{tabular}%
}
\caption{Effectiveness of different distillation strategies used in our method. '+' denotes the add operation.}
\label{tab:distill}
\end{table}

\begin{figure}[!t]
  \centering
  \includegraphics[width=\linewidth]{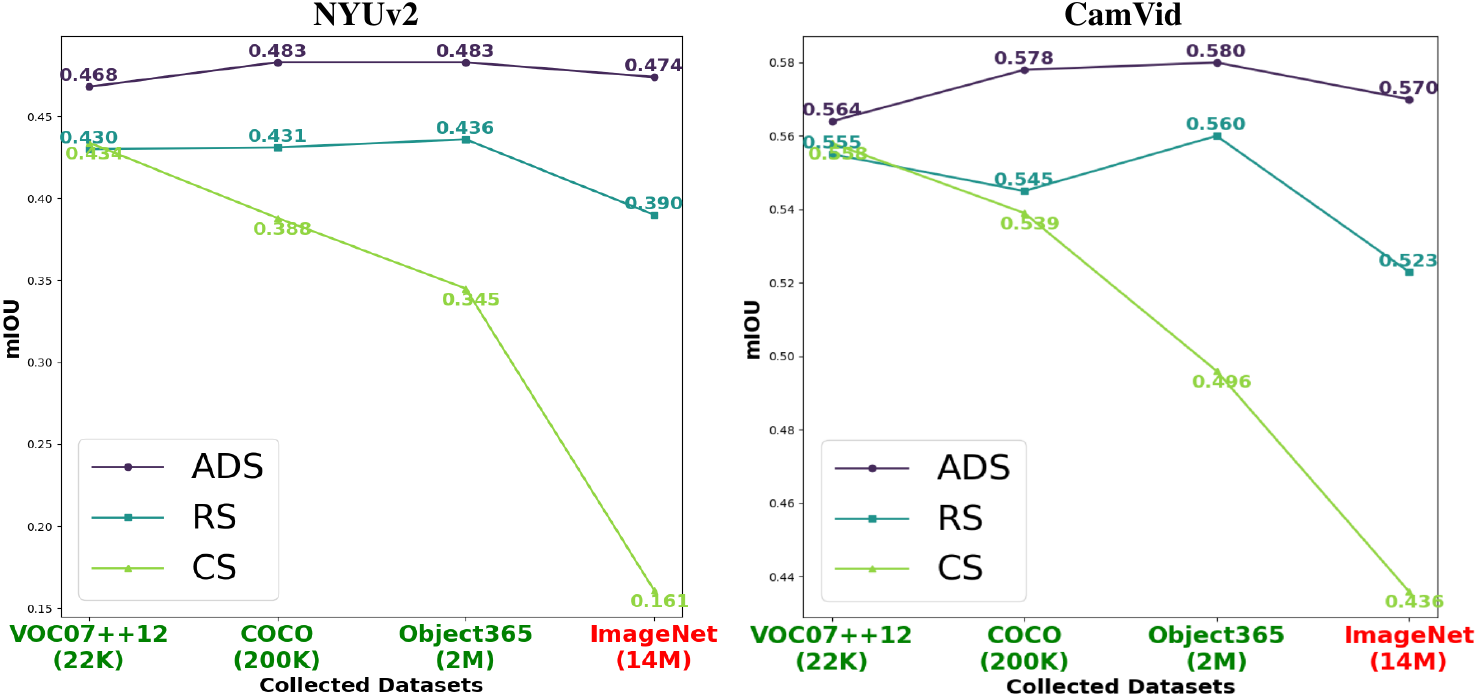}
  \caption{The performance of different sampling methods on different collected datasets. The x-axis represents different datasets, with the corresponding dataset size in parentheses. The color represents the task complexity of the dataset, with green for complex tasks and red for simple tasks.}
  \label{fig:further}
\end{figure}

\noindent\textbf{Analysis of Collected Datasets.} We extend our evaluation to additional open-world datasets (Object365~\cite{shao2019object365}, VOC07++12~\cite{everingham2010voc}). Fig.~\ref{fig:further} reveals two key insights: (1) confidence-based methods degrade significantly with larger datasets while ADS maintains robust performance, and (2) ADS's effectiveness depends on dataset characteristics. While larger datasets enable ADS to select more distribution-aligned samples, this improvement requires the dataset's task characteristics to closely match segmentation requirements for sufficient information richness. For instance, despite its large scale, ImageNet's simplicity limits its effectiveness for segmentation tasks.

To further study the inherent characteristics of collected datasets, we use image information entropy to evaluate each collected dataset. Table~\ref{tab:entropy} shows that compared to ImageNet designed for classification tasks, other collected datasets consistently demonstrate higher information richness and more stable distributions. However, even information-rich datasets like VOC07++12 underperform when their scale is insufficient. These findings demonstrate that effective dataset selection must simultaneously consider both scale and information richness to optimize knowledge transfer.

\begin{table}[h]
\centering
\resizebox{0.475\textwidth}{!}{
\begin{tabular}{@{}cccccc@{}}
\toprule
          & Mean  & Median & Variance & CamVid & NYUv2 \\ \midrule
ImageNet  & 7.179 & 7.399  & 0.560    & 0.570  & 0.474 \\
VOC07++12 & 7.300 & 7.453  & 0.328    & 0.564(-0.006)  & 0.468(-0.006) \\
COCO      & 7.348 & 7.482  & 0.260    & 0.578(+0.008)  & 0.483(+0.009) \\
Object365 & 7.369 & 7.502  & 0.248    & 0.580(+0.010)  & 0.483(+0.009) \\ \bottomrule
\end{tabular}
}
\caption{Image information entropy statistics and ADS performance across different collected datasets. Mean, median, and variance represent the statistical characteristics of image entropy over the complete dataset. We use the performance of sampling on ImageNet as baseline.}
\label{tab:entropy}
\end{table}